\pdfoutput=1

\documentclass[11pt]{article}

\usepackage[final]{mtsummit25}
 \usepackage{copyright}

\usepackage{times}
\usepackage{latexsym}

\usepackage[T1]{fontenc}

\usepackage[utf8]{inputenc}

\usepackage{microtype}

\usepackage{inconsolata}

\usepackage{graphicx}

\title{Exploring the Feasibility of Multilingual Grammatical Error Correction with a Single LLM up to 9B parameters: \\A Comparative Study of 17 Models}

\author{
  \textbf{Dawid Wisniewski\textsuperscript{1,2}},
  \textbf{Antoni Solarski\textsuperscript{1,3}},
  \textbf{Artur Nowakowski\textsuperscript{1,3}},
\\
  \textsuperscript{1}Laniqo.com,
  \textsuperscript{2}Poznan University of Technology, Poland,
  \textsuperscript{3}Adam Mickiewicz University, Poland
\\
  \small{
    \textbf{Correspondence:} \href{mailto:email@domain}{dawid.wisniewski@laniqo.com}
  }
}

\begin{document}
\maketitle
\begin{abstract}
Recent language models can successfully solve various language-related tasks, and many understand inputs stated in different languages. In this paper, we explore the performance of 17 popular models used to correct grammatical issues in texts stated in English, German, Italian, and Swedish when using a single model to correct texts in all those languages. We analyze the outputs generated by these models, focusing on decreasing the number of grammatical errors while keeping the changes small. The conclusions drawn help us understand what problems occur among those models and which models can be recommended for multilingual grammatical error correction tasks. We list six models that improve grammatical correctness in all four languages and show that Gemma 9B is currently the best performing one for the languages considered\footnote{Project supported by grant no. 0311/SBAD/0763 - Mloda Kadra financed by Poznan University of Technology.
}.
\end{abstract}

\section{Introduction}
Grammatical error correction (GEC) is one of the most practical tasks in the Natural Language Processing (NLP) field. Being able to use computers to detect and fix grammatical errors and spelling mistakes is especially beneficial for language learners and professional writers. Recent years have shown great promise in using Large Language Models (LLMs) for various NLP-related tasks, including machine translation, text generation, or text classification. These models, trained on vast amounts of data, learn to predict the most probable continuation of a given sequence. Assuming large corpora are used to train these models, the models should encounter instantiations of various tasks, including questions to be answered followed by their answers, texts expressed in one language followed by their translations, long fragments of texts followed by their summarizations, or grammatically incorrect sentences fixed by some experts as part of the text's continuation. All of these lead to the increasing ability of LLMs to address human requests.

Recently published LLMs are bigger and trained on more data, which enables them to solve more sophisticated tasks with better results. Even though LLMs are currently suboptimal in some tasks, it seems that soon they may become the dominant solution for most NLP-related problems~\cite{DBLP:journals/corr/abs-2001-08361}. However, due to LLMs' large sizes and high computational costs, researchers focus on training smaller models of quality similar to bigger ones~\cite{DBLP:journals/computer/Shan24}.

In this paper, we aim to explore the abilities of LLMs to solve the GEC task when dealing with several languages at once, including three highly popular ones: English (EN), German (DE), and Italian (IT), as well as a less popular -- Swedish (SV). Having one common model for multiple languages may be beneficial in various ways. Using specialized models for each language requires a lot of storage space, which increases the costs of products and hinders the synergy between languages, which is frequently observed in real life. The energy consumption related to using multiple models increases the costs of serving models and impacts our natural environment. To mitigate these issues, smaller models that can be run locally and are more environmentally sustainable may be considered~\cite{small-llms}. These smaller models are easier to control, as they can be loaded, analyzed, and fine-tuned on personal computers, so here, we focus on using single LLMs of moderate size (up to 9B parameters) as they can be loaded on consumer-grade GPUs.

In this paper, we aim to address the following research questions:\\\textbf{RQ1}: Which model is the best for multilingual grammar correction considering English, German, Italian, and Swedish?
\\\textbf{RQ2}: Do models preserve the original input when no errors are present?
\\\textbf{RQ3}: Which type of prompt is more effective: short and general, or longer and more specific?

\section{Related works}

The research on the GEC problems has experienced substantial progress in the last years. Recent surveys highlight a shift from traditional rule-based methods, statistical classifiers, and statistical machine translation techniques to neural machine translation, emphasizing the superior performance of neural approaches~\cite{survey-wang, survey-bryant}. However, there are challenges associated with the predominant use of supervised learning in GEC, primarily due to the necessity for annotated datasets. Researchers in the field have reported issues such as annotation inconsistency, human error, and a scarcity of data for languages other than English~\cite{survey-bryant}. In this context, leveraging LLM-based GEC with zero or a few examples emerges as a promising solution. Aside from unsupervised approaches, other significant challenges in GEC include multilingualism, low-resource GEC, and the evaluation~\cite{survey-wang, survey-bryant}.

Here, we focus on three main areas relevant to our research: the application of LLMs in GEC, multilingual GEC, and the evaluation of GEC systems.

\subsection{Large language models for GEC}

In the domain of GEC, LLMs have been utilized in several innovative ways. For instance, LLMs were employed as evaluators of GEC systems~\cite{2024largelanguagemodelsstateoftheart}. Large-scale models like GPT-4~\cite{DBLP:journals/corr/abs-2303-08774} have achieved state-of-the-art results in GEC evaluation, demonstrating a higher correlation with human judgments than other methods~\cite{2024largelanguagemodelsstateoftheart}. Also, it was proposed to integrate a language model as a grammatical error detection module, thereby enhancing the overall performance of GEC systems~\cite{lmcritic}. Nevertheless, LLMs were explored as standalone GEC systems for English~\cite{2024-prompting}. By employing effective prompting techniques, the authors evaluated seven open-source models and three commercial models. Their findings suggest that LLMs can outperform supervised GEC models on benchmarks annotated with fluency corrections. Furthermore, the study shows that zero-shot generation can be as effective as few-shot approaches. A related study was also conducted for Swedish~\cite{2023-swedish-llm}, where the properly prompted GPT-3 model heavily outperformed other GEC systems. Similar research, for the same model, was also conducted for English \cite{gpt3grammatical}, where authors focused on the controllability aspect of the prompt-based approach. This versatility highlights the significant potential of LLMs to advance GEC methodologies. 

\subsection{Multilingual GEC}

As multilingualism emerges as a promising direction in GEC, researchers are actively exploring methods to develop and enhance GEC systems for low-resource languages. For instance, certain strategies were proposed for training and fine-tuning GEC models to create a single system capable of handling multiple languages~\cite{simplerecipemultilingualgrammatical, pajak-multilingual}. Similarly, transfer learning was employed to leverage models trained on high-resource languages \cite{2020-cross-lingual}. Many LLMs demonstrate at least some degree of multilingual capability, making it logical to explore their potential in the context of multilingual GEC. For instance, the large commercial GPT-3.5 model was examined for its performance in GEC across various languages, yielding promising results. However, the human evaluation revealed that the model encounters difficulties with specific types of errors~\cite{2024-gpt-3.5}. Collectively, this collection of research demonstrates the feasibility of developing a unified model capable of performing GEC across various languages, including those with limited resources. However, different LLMs, particularly smaller ones, haven't yet been investigated in this context.

\subsection{Evaluations}

In GEC evaluation, metrics can be broadly categorized into two main types: reference-based and reference-less~\cite{survey-bryant, survey-wang}. While different reference-based metrics are well-studied and widely used, the focus is still on developing improved metrics, particularly those that do not rely on references. It was noted that the existing evaluation methods might stem the progress in the field, as they are tightly coupled with gold-standard references~\cite{rozovskaya-2021-good}.

Reference-less metrics for assessing grammatical correctness are very rich. Some of the popular choices use e-rater~\cite{erater} and LanguageTool~\cite{languagetool} to calculate grammatical correctness scores. Additionally, an important aspect of GEC systems is the preservation of meaning, for which the $US_{IM}$~\cite{USim} or BERTScore~\cite{bert-score} metrics can be employed. Therefore, further research in GEC evaluation, particularly in the area of semantic faithfulness (meaning preservation), is postulated~\cite{survey-wang}. Other aspects are text fluency, for which e.g., the GLEU~\cite{DBLP:conf/acl/MuttonDWD07} metric is frequently used, or ensuring that corrections are made using minimal changes rather than reformulations of the whole sentence, which can be analyzed using e.g., Levenshtein distance~\cite{DBLP:journals/coling/Keselj09}.

\section{Dataset}
We use the MultiGED dataset~\cite{volodina2023multiged} for our experiments. Originally, MultiGED was proposed in 2023 to evaluate the ability of AI systems to identify grammatically incorrect tokens in five languages: English, German, Italian, Czech, and Swedish. 

The dataset is a compilation of various datasets that include FCE for English~\cite{DBLP:conf/acl/YannakoudakisBM11}, or Falko-MERLIN for German~\cite{DBLP:conf/aclnut/Boyd18}. As the original dataset represents a binary classification task, where each token in a sentence is classified as correct or not, we preprocessed the dataset using Moses detokenizer~\cite{koehn-etal-2007-moses}, to reconstruct entire sentences annotated with the information whether the whole sentence is grammatically correct or not. If any token in the sentence was annotated as incorrect, the full sentence was tagged as incorrect. 

There is no golden-standard corrected candidate for a given sentence provided, as the MultiGED dataset is a token classification task. However, following research of~\cite{rozovskaya-2021-good}, our goal is to analyze the GEC problem from a broader perspective trying to mitigate annotator bias that may be observed in golden standard corrections. 

To achieve this, we use LanguageTool as one of the key tools for scoring the models. As it does not support Czech, we focus on English, German, Italian and Swedish only. To evaluate a wide range of LLMs, we selected the MultiGED's dev set to limit the costs of experiments. This step reduced inference computational power needs, leaving relatively large number of examples for each language considered. The summary of the processed dataset used for further experiments is provided in Table~\ref{tab:multiged_processed}.
\begin{table}
  \small
  \centering
  \begin{tabular}{lcccc}
    \hline
    \textbf{Language} & \textbf{Total sents} & \textbf{Correct} & \textbf{Tokens per sent} \\
    \hline
    English & 2191 & 906 & 15.9 (+/- 10.97) \\
    German & 2503 & 619 &  15.81 (+/- 9.46) \\
    Italian & 758 & 268 & 11.98 (+/- 7.61)\\
    Swedish & 911 & 199 & 17.25 (+/- 11.43)\\ \hline 
    TOTAL & 6363 & 1992 & 15.59 (+/- 10.21) \\ \hline
  \end{tabular}
  \caption{Dataset summary. \textbf{Total sents} column represents the number of sentences for a given language, \textbf{Correct} represents the number of sentences marked as grammatically correct, and \textbf{Tokens per sent} represents the average number of tokens in a sentence, with standard deviation provided.} 
  \label{tab:multiged_processed}
\end{table}

\section{Models and Methodology}

\subsection{Models selection}
We searched for language models meeting the following criteria: (i) They should fit popular consumer GPUs, thus we set the size limit to 9B parameters, (ii) They, or their base models, should be accompanied by a research paper, (iii) We prefer instruction-following models if present. 

As a result, we collected 17 LLMs listed in Table~\ref{tab:llms}. From these, 8 have 7B parameters, 2 have 8B parameters, 3 have 9B parameters, and 4 are smaller than 4B parameters. All models but XGLM and Bloom are instruction-following ones. We decided to consider XGLM and Bloom anyway as their multilingual abilities are strongly underlined in their research papers. Although Karen-strict and Karen-creative are not accompanied by research papers, they are fine-tuned Mistral models that were trained directly to solve the grammatical error correction task. 

\subsection{Model querying}
\label{sec:prompts}

Every model was queried using three user prompts, which were translated into the target language for inputs other than English (DE, IT, SV): 

\textbf{P1}: \textit{Edit the following text for spelling and grammar mistakes:}

\textbf{P2}: \textit{Edit the following text for spelling and grammar mistakes, return only the corrected text:}

\textbf{P3}: \textit{Edit the following text for spelling and grammar mistakes, make minimal changes, and return only the corrected text. If the text is already correct, return it without any explanations:}.


For generation, we use \texttt{transformers} library, version 4.42.4. We used each model's \texttt{.generate()} function, setting the following parameters:

- \texttt{renormalize\_logits=False}, 

- \texttt{do\_sample=True}, 

- \texttt{use\_cache=True}, 

- \texttt{max\_new\_tokens=256}, 

- \texttt{repetition\_penalty=1.18},

- \texttt{top\_k=40}, 

- \texttt{top\_p=0.1}. 

The parameter selection is inspired by Karen~\footnote{\url{https://huggingface.co/FPHam/Karen_TheEditor_V2_STRICT_Mistral_7B}}, which is the only model (considered) fine-tuned for GEC. It provides these values as suggestions. We share those values among all models, and leave other generation parameters default.


\begin{table*}
  \centering
  \small
  \begin{tabular}{ll}
    \hline
    \textbf{Name} & \textbf{Huggingface ID} \\ 
    \hline
    Aya (8B)~\cite{aryabumi2024aya} & \verb|CohereForAI/aya-23-8B|  \\ \hline 
    Bloom (7B)~\cite{le2023bloom} & \verb|bigscience/bloom-7b1| \\ \hline
    EuroLLM (1.7B)~\cite{DBLP:journals/corr/abs-2409-16235} & \verb|utter-project/EuroLLM-1.7B-Instruct|\\ \hline 
    EuroLLM (9B)~\cite{DBLP:journals/corr/abs-2409-16235} & \verb|utter-project/EuroLLM-9B-Instruct|\\ \hline 
    Gemma 2 2B (2B)\cite{team2024gemma} & \verb|google/gemma-2-2b-it| \\\hline
    Gemma 2 9B (9B)~\cite{team2024gemma} & \verb|google/gemma-2-9b-it|\\ \hline
    Karen-creative (7B)~\cite{jiang2023mistral} & \verb|FPHam/Karen_TheEditor_V2_CREATIVE_Mistral_7B| \\ \hline  
    Karen-strict (7B)~\cite{jiang2023mistral} & \verb|FPHam/Karen_TheEditor_V2_STRICT_Mistral_7B| \\ \hline
    Llama 3.1 (8B)~\cite{touvron2023llama} & \verb|meta-llama/Meta-Llama-3.1-8B-Instruct| \\ \hline
    Mistral (7B)~\cite{jiang2023mistral} & \verb|mistralai/Mistral-7B-Instruct-v0.3|\\ \hline
    OpenChat 3.5 (7B)~\cite{wang2023openchat} & \verb|openchat/openchat-3.5-0106|\\ \hline
    Phi-3 (3.8B)~\cite{abdin2024phi} & \verb|microsoft/Phi-3-mini-4k-instruct|  \\ \hline
    Qwen 2.5 (7B)~\cite{yang2024qwen2} & \verb|Qwen/Qwen2.5-7B-Instruct|\\ \hline
    SmolLM (1.7B)~\cite{allal2024SmolLM}& \verb|HuggingFaceTB/SmolLM-1.7B-Instruct|  \\ \hline
    TowerLLM (7B)~\cite{tower_llm_2024} & \verb|Unbabel/TowerInstruct-7B-v0.2|  \\ \hline
    XGLM (7.5B)~\cite{lin2021few} & \verb|facebook/xglm-7.5B| \\ \hline
    Yi (9B)~\cite{young2024yi} & \verb|01-ai/Yi-1.5-9B-Chat| \\ \hline

  \end{tabular}
  \caption{Models selected for the experiment}
  \label{tab:llms}
\end{table*}

\subsection{Analyzed characteristics}
\label{sec:metrics}
A good GEC system should meet several criteria: (i) it should fix grammatical issues in the text, (ii) it should preserve the meaning of the original text, (iii) it should make possibly small changes to the input text. To gain a better understanding of the LLMs' outputs, we introduced two additional criteria: (iv) it should detect when a given text is correct and should not be changed, and (v) it should preserve the language of the original text. These criteria highlight the inherent dilemma in GEC systems: the trade-off between prioritizing corrections and preserving the original text. The focus on meaning and language preservation stems from the fact that LLMs may generate texts that are not corrected texts (e.g., \textit{There are no errors in this text.}) or may fall back to a different language (e.g., English) if they do not understand a given one. To address these requirements, the following metrics are used, which are calculated for each sentence and then averaged over all examples in a given language: 


\paragraph{Req. 1: Grammatical correctness}

To evaluate language correctness, we use the Python wrapper for LanguageTool~\footnote{\url{https://languagetool.org/}} (LT), which provides us with information about the list of grammatical errors found in a given text. We use information on the number of errors to evaluate an input sentence $s$: $$correctness(s) = \frac{1}{1 + num\_errors(s)}$$ This metric ranges between 0.00 (really bad quality of output) and 1.0 (no grammatical errors found by LT). Good models should increase the value of that metric after correction. The decision to use this metric over other ones is due to the multilingual scenario -- LT supports all 4 languages considered, while e.g., e-rater does not.




\paragraph{Req 2: Semantic similarity} To assess the semantic similarity between the uncorrected input sentence $s_i$ and the corrected output sentence $s_o$ and detect situations where the text generated is not a correction (e.g., \textit{No errors found}), we calculate three auxiliary metrics, namely: (i) BERTScore~\cite{bert-score} similarity calculated using a multilingual BERT~\footnote{google-bert/bert-base-multilingual-cased}~\cite{DBLP:journals/corr/abs-1810-04805}, (ii) BLEURT~\cite{sellam-etal-2020-bleurt} similarity calculated using a multilingual model~\footnote{BLEURT-20-D12}, (iii) SentenceBERT~\cite{DBLP:conf/emnlp/ReimersG19} similarity calculated as a cosine similarity between representations of $s_i$ and $s_o$ generated using a multilingual SentenceBERT model~\footnote{sentence-transformers/paraphrase-multilingual-MiniLM-L12-v2}.

We decided to use three metrics instead of one to make the results more robust and less biased towards one model.


\paragraph{Req 3: Syntactic similarity}
Apart from reducing the number of errors and preserving meaning, ideally, the model should apply minimal changes to the input text. Similarly to semantic similarity, also here, we use three auxiliary metrics:  
(i) Levenshtein (edit) score $e()$, which is a transformed edit distance between two sentences $s_i$ and $s_o$. Levenshtein (edit) distance defines how many changes to the source text $s_i$ should be applied to obtain the target text $s_o$. The score is calculated as follows: $$e(s_i, s_o) = \frac{1}{1 + edit\_distance(s_i, s_o)}$$
(ii) GLEU score $g()$, which is one of the most popular metrics for GEC~\cite{DBLP:conf/acl/MuttonDWD07}. GLEU measures the overlap between the tokens of a hypothesis (here, generated sentence $s_o$) and a set of references (here, input sentence $s_i$). The score is calculated using NLTK's~\cite{bird2009natural} \texttt{sentence\_gleu} function with default parameters.
(iii) Length difference $d()$, which calculates the difference between $s_i$ and $s_o$ in terms of token numbers. The score is calculated using the following equation, where $cnt()$ returns the number of elements, tokens are generated by $tok()$ function using \texttt{word\_tokenize()} method from NLTK, and $max()$ returns the maximum value:

    $$d(s_i, s_o) = 1 - \frac{|cnt(tok(s_i)) - cnt(tok(s_o))|}{max(cnt(tok(s_i)),cnt(tok(s_o)))}$$

Although one may expect that tokens may be added or removed (e.g., punctuation marks) after correction, this metric is most sensitive to producing completely different text (e.g., empty output). 


    

\paragraph{Req 4: Keeping correct sentences unchanged} Besides correcting errors, LLMs should be able to identify cases, where an input sentence $s_i$ is grammatically correct. The expected behavior in that case is to keep $s_i$ unchanged and return it as the output sentence $s_o$. Having information about the correctness of sentences from MultiGED, we define true positives, false positives, and false negatives as: the number of examples, where $s_i = s_o$, when $s_i$ is marked correct, the number of examples, where $s_i = s_o$, when $s_i$ is marked incorrect, and the number of examples, where $s_i \neq s_o$ and $s_i$ is marked correct, respectively. Then, we calculate the $F_1$ score based on these.


\paragraph{Req 5: Language drift} Some LLMs have tendencies to produce outputs in a different language than the desired one~\cite{DBLP:journals/corr/abs-2406-20052}. This behavior is most frequently observed when a given text expressed in a language other than English is transformed into a text in English without explicitly asking for this kind of switch. For this reason, having an input sentence $s_i$ expressed in language $l$, that is corrected using a given LLM into a new text $s_o$, we use the Language Identification tool called LID~\footnote{https://huggingface.co/facebook/fasttext-language-identification}. LID supports 217 languages and is based on Fasttext~\cite{bojanowski2016enriching}. Knowing what language $l$ a given text $s_i$ is expressed in, we report the probability change:

$$
    drift(s_o, s_i, l) = P(l|s_o) - P(l|s_i)
$$

Since the corrected text should be more probable to be observed, we expect this metric to be $\ge$ 0.0. Values well below 0 mean that the LID tool is more confused about the language after correction, which is not a desired behavior. 

\section{Results}

\begin{table*}
  \tiny
  \centering
  \begin{tabular}{lccccccccc}
    \hline
    \textbf{Prompt [lang]} & \textbf{LT$\uparrow$} & \textbf{BERT Score$\uparrow$} & 
    \textbf{SentenceBERT$\uparrow$} & \textbf{BLEURT$\uparrow$} & \textbf{Levenshtein$\uparrow$} & \textbf{Length diff$\uparrow$} & \textbf{GLEU$\uparrow$} & \textbf{Language drift$\uparrow$} & \textbf{Correct ($F_1$)$\uparrow$} \\
    \hline

P1[EN] & 0.740 & 0.750 & 0.651 & 0.522 & 0.092 & 0.486 & 0.383 & 0.049 & 0.124 \\
P2[EN] & 0.804 & 0.820 & \textbf{0.739} & 0.641 & 0.169 & \textbf{0.694} & 0.570 & \textbf{0.035} & 0.278 \\
P3[EN] & \textbf{0.807} & \textbf{0.824} & 0.735 & \textbf{0.647} & \textbf{0.197} & 0.689 & \textbf{0.577} & \textbf{0.035} & \textbf{0.342} \\ \hline 
P1[DE] & 0.707 & 0.739 & 0.659 & 0.446 & 0.061 & 0.461 & 0.363 & \textbf{-0.180} & 0.103 \\
P2[DE] & 0.790 & 0.794 & 0.735 & 0.544 & 0.090 & 0.641 & 0.509 & -0.183 & 0.174 \\
P3[DE] & \textbf{0.811} & \textbf{0.805} & \textbf{0.738} & \textbf{0.560} & \textbf{0.110} & \textbf{0.677} & \textbf{0.534} & -0.212 & \textbf{0.244} \\ \hline 
P1[IT] & 0.497 & 0.719 & 0.624 & 0.375 & 0.056 & 0.443 & 0.298 & \textbf{-0.242} & 0.065 \\
P2[IT] & 0.601 & 0.775 & 0.698 & 0.502 & 0.111 & 0.615 & 0.452 & -0.244 & 0.199 \\
P3[IT] & \textbf{0.638} & \textbf{0.787} & \textbf{0.711} & \textbf{0.532} & \textbf{0.131} & \textbf{0.658} & \textbf{0.487} & -0.262 & \textbf{0.249} \\ \hline 
P1[SV] & 0.420 & 0.728 & 0.646 & 0.395 & 0.048 & 0.444 & 0.306 & -0.284 & 0.081 \\
P2[SV] & 0.524 & 0.792 & 0.716 & 0.512 & 0.094 & 0.655 & 0.486 & -0.286 & 0.187 \\
P3[SV] & \textbf{0.552} & \textbf{0.804} & \textbf{0.725} & \textbf{0.539 }& \textbf{0.113} & \textbf{0.682} & \textbf{0.518} & \textbf{-0.283} & \textbf{0.248} \\ \hline

  \end{tabular}
  \caption{Prompt selection vs. metrics for each language. For each language and prompt, a given row represents scores averaged over all 17 models considered. Prompt identifiers are the same as introduced in Section~\ref{sec:prompts}.}
  \label{tab:prompts}
\end{table*}

\begin{table*}
  \tiny
  \centering
  \begin{tabular}{lccccccccc}
    \hline
    \textbf{Model} & \textbf{LT$\uparrow$} & \textbf{BERT Score$\uparrow$} & 
    \textbf{SentenceBERT$\uparrow$} & \textbf{BLEURT$\uparrow$} & \textbf{Levenshtein$\uparrow$} & \textbf{Length diff$\uparrow$} & \textbf{GLEU$\uparrow$} & \textbf{Language drift$\uparrow$} & \textbf{Correct ($F_1$)$\uparrow$} \\
    \hline
Aya & 0.900 & 0.920 & 0.935 & 0.772 & 0.259 & \textbf{0.928 (3)} & \textbf{0.797 (3)} & 0.015 & \textbf{0.481 (3)} \\
BLOOM & 0.184 & 0.516 & 0.028 & 0.249 & 0.001 & 0.056 & 0.032 & -0.693 & 0.000 \\
EuroLLM (1.7B) & 0.912 & 0.849 & 0.848 & 0.639 & 0.119 & 0.854 & 0.583 & 0.011 & 0.223 \\
EuroLLM (9B) & 0.915 & 0.888 & 0.902 & 0.712 & 0.218 & 0.790 & 0.702 & \textbf{0.029 (2)} & 0.437 \\
Gemma (2B) & \textbf{0.940 (2)} & \textbf{0.920 (3)} & \textbf{0.941 (3)} & 0.755 & 0.121 & 0.926 & 0.774 & 0.012 & 0.442 \\
Gemma (9B) & \textbf{0.948 (1)} & \textbf{0.937 (1)} & \textbf{0.954 (1)} & \textbf{0.774 (3)}  & 0.136 & \textbf{0.942 (1)} & \textbf{0.814 (2)} & 0.017 & \textbf{0.560 (1)} \\
Karen (creative) & 0.705 & 0.895 & 0.933 & 0.643 & \textbf{0.276 (2)} & 0.924 & 0.687 & -0.297 & 0.457 \\
Karen (strict) & 0.662 & 0.879 & 0.913 & 0.575 & \textbf{0.268 (3)} & 0.906 & 0.622 & -0.432 & 0.463 \\
Llama 3.1 & \textbf{0.937 (3)} & 0.887 & 0.894 & 0.709 & 0.148 & 0.853 & 0.700 & \textbf{0.032 (1)} & 0.255 \\
Mistral & 0.710 & 0.810 & 0.847 & 0.532 & 0.044 & 0.655 & 0.489 & -0.151 & 0.022 \\
OpenChat & 0.934 & 0.919 & 0.932 & \textbf{0.775 (2)} & 0.242 & 0.925 & 0.793 & \textbf{0.018 (3)} & 0.420 \\
Phi & 0.428 & 0.655 & 0.533 & 0.285 & 0.006 & 0.230 & 0.142 & -0.065 & 0.002 \\
Qwen 2.5 & 0.896 & \textbf{0.935 (2)} & \textbf{0.952 (2)} & \textbf{0.805 (1)} & \textbf{0.298 (1)} & \textbf{0.939 (2)} & \textbf{0.845 (1)} & 0.008 & \textbf{0.545 (2)} \\
SmolLM & 0.117 & 0.536 & 0.060 & 0.253 & 0.001 & 0.065 & 0.031 & -0.695 & 0.000 \\
TowerLLM & 0.653 & 0.832 & 0.840 & 0.503 & 0.141 & 0.804 & 0.500 & -0.384 & 0.251 \\
XGLM & 0.366 & 0.513 & 0.095 & 0.168 & 0.007 & 0.167 & 0.044 & -0.508 & 0.000 \\
Yi & 0.729 & 0.788 & 0.754 & 0.532 & 0.058 & 0.536 & 0.436 & 0.015 & 0.047 \\ \hline

  \end{tabular}
  \caption{Scores macro-averaged over languages (EN, DE, IT, SV).}
  \label{tab:scores_average}
\end{table*}

\begin{table*}
  \tiny
  \centering
  \begin{tabular}{l||c||cccccc}
    \hline
    \textbf{Model} & \textbf{Aggregated rank} & \textbf{Rank EN} & \textbf{Rank DE} & \textbf{Rank IT} & \textbf{Rank SV} & \textbf{Supports all langs} & \textbf{Improves LT on} \\
    \hline
    \textbf{Gemma (9B)} & 1 & 4 & 6 & 1 & 1 & YES & EN, DE, IT, SV\\
    Qwen 2.5 & 2 & 2 & 1 & 3 & 2 & YES & EN, DE, IT\\
    Aya & 3 & 7 & 2 & 2 & 5 & YES & EN, DE, IT\\
    \textbf{Gemma (2B)} & 4 & 7 & 4 & 4 & 3 & YES & EN, DE, IT, SV\\
    \textbf{OpenChat} & 5 & 5 & 2 & 4 & 4 & YES & EN, DE, IT, SV\\
    \textbf{EuroLLM (9B)} & 6 & 6 & 8 & 7 & 7 & YES & EN, DE, IT, SV\\
    Karen (creative) & 6 & 2 & 5 & 11 & 5 & NO & EN, DE\\
    \textbf{Llama 3.1} & 8 & 9 & 7 & 9 & 8 & YES & EN, DE, IT, SV\\
    Karen (strict) & 9 & 1 & 8 & 7 & 10 & NO & EN, DE\\
    \textbf{EuroLLM (1.7B)} & 10 & 12 & 11 & 6 & 9 & YES & EN, DE, IT, SV \\
    TowerLLM & 11 & 11 & 10 & 10 & 14 & NO & EN, DE \\
    Mistral & 12 & 10 & 12 & 13 & 12 & NO & EN, DE\\
    Yi & 13 & 13 & 13 & 12 & 11 & YES & EN, DE\\
    Phi & 14 & 14 & 14 & 14 & 13 & NO & --\\
    BLOOM & 15 & 15 & 16 & 16 & 15 & NO & --\\
    XGLM & 15 & 16 & 15 & 15 & 16 & NO & --\\
    SmolLM & 17 & 16 & 16 & 17 & 17 & NO & --\\
    \hline

  \end{tabular}
  \caption{Aggregated ranks with ties represented as the same positions. Models in bold support all languages and improve correctness (LT) metric on each language. Since Aya, Qwen 2.5, and Karen are very good on some languages (Karen is top-scored on English, Aya is second on Italian and German, and Qwen 2.5 is top-scored for German), they outperform some other models marked in bold, which support all languages but with worse quality.}
  \label{tab:aggregated_scores}
\end{table*}

We calculated values for all the metrics introduced in Section~\ref{sec:metrics} using all the prompts described in Section~\ref{sec:prompts} for all 17 LLMs considered. Then, we selected the best performing prompt, and used it to verify which LLMs support all the languages, and which LLMs work best for multilingual scenarios as well as for each language separately.


\paragraph{Prompt selection}
For each metric, each language, and each prompt \textbf{P1}, \textbf{P2}, \textbf{P3}, we calculated the average metric value over all models considered. While the detailed scores can be seen in Table~\ref{tab:prompts}, we observe that the best performing prompt is prompt \textbf{P3}, which is the longest and most concrete one. This prompt was selected as the best one in 32/36 scenarios considered.

The biggest gain of using the last prompt is seen in the $F_1$ metric, which checks how well LLMs keep correct sentences unchanged. This observation agrees with the intuition -- in prompt \textbf{P3}, we explicitly tell those models not to change the input in the last scenario. The significant increase of $F_1$ score for each language shows that the model understands this kind of query. In the subsequent analyses, prompt \textbf{P3} is used.

\paragraph{Language support analysis}

Several LLMs struggle with handling languages other than English, often producing outputs (or considerable fragments of outputs) in English. Models like Bloom and SmolLM exhibit the highest drifts, converting a significant portion of German, Italian, and Swedish texts into English. Other models, such as XGLM, Karen-creative, Karen-strict, TowerLLM, Mistral, and Phi-3, also show considerable drift, with each producing outputs in a different language in at least a quarter of the cases. Consequently, these models are not considered effective for multilingual grammatical error correction. Detailed per-language results can be found in Appendix, in Tables~\ref{tab:appendix_scores_en},~\ref{tab:appendix_scores_de},~\ref{tab:appendix_scores_it},~\ref{tab:appendix_scores_sv}. The aggregated scores can be found in Table~\ref{tab:scores_average}.


\begin{table*}
  \tiny
  \centering
  \begin{tabular}{lccccccccc}
    \hline
    \textbf{Model} & \textbf{LT$\uparrow$} & \textbf{BERT Score$\uparrow$} & 
    \textbf{SentenceBERT$\uparrow$} & \textbf{BLEURT$\uparrow$} & \textbf{Levenshtein$\uparrow$} & \textbf{Length diff$\uparrow$} & \textbf{GLEU$\uparrow$} &  \textbf{Language drift$\uparrow$} &  \textbf{Correct ($F_1$$)\uparrow$} \\
    \hline
Aya & 7 & 4 & 4 & 4 & 4 & \textbf{3} & \textbf{3} & 6 & \textbf{3} \\
BLOOM & 16 & 16 & 17 & 16 & 16 & 17 & 16 & 16 & 15 \\
EuroLLM (1.7B) & 6 & 10 & 10 & 9 & 11 & 8 & 10 & 8 & 11 \\
EuroLLM (9B) & 5 & 7 & 8 & 6 & 6 & 11 & 6 & \textbf{2} & 7 \\
Gemma (2B) & \textbf{2} & \textbf{3} & \textbf{3} & 5 & 10 & 4 & 5 & 7 & 6 \\
Gemma (9B) & \textbf{1} & \textbf{1} & \textbf{1} & \textbf{3} & 9 & \textbf{1} & \textbf{2} & 4 & \textbf{1} \\
Karen (creative) & 11 & 6 & 5 & 8 & \textbf{2} & 6 & 8 & 12 & 5 \\
Karen (strict) & 12 & 9 & 7 & 10 & \textbf{3} & 7 & 9 & 14 & 4 \\
Llama 3.1 & \textbf{3} & 8 & 9 & 7 & 7 & 9 & 7 & \textbf{1} & 9 \\
Mistral & 10 & 12 & 11 & 11 & 13 & 12 & 12 & 11 & 13 \\
OpenChat & 4 & 5 & 6 & \textbf{2} & 5 & 5 & 4 & \textbf{3} & 8 \\
Phi & 14 & 14 & 14 & 14 & 15 & 14 & 14 & 10 & 14 \\
Qwen 2.5 & 8 & \textbf{2} & \textbf{2} & \textbf{1} & \textbf{1} & \textbf{2} & \textbf{1} & 9 & \textbf{2} \\
SmolLM & 17 & 15 & 16 & 15 & 17 & 16 & 17 & 17 & 16 \\
TowerLLM & 13 & 11 & 12 & 13 & 8 & 10 & 11 & 13 & 10 \\
XGLM & 15 & 17 & 15 & 17 & 14 & 15 & 15 & 15 & 17 \\
Yi & 9 & 13 & 13 & 12 & 12 & 13 & 13 & 5 & 12 \\
    \hline

  \end{tabular}
  \caption{Ranks generated based on Table~\ref{tab:scores_average}}
  \label{tab:ranks_average}. 
\end{table*}

Out of 17 LLMs considered, 9 of them (Aya, EuroLLM 1.7B, EuroLLM 9B, Gemma 2B and 9B, LLama  3.1, Openchat 3.5, Qwen 2.5, and Yi) produce outputs in all languages considered in a vast majority of cases. Mistral and Phi-3 fail to produce texts in one language - Italian, while the remaining models have problems with at least two languages.  Since not all LLMs support all languages, we repeated the prompt quality analysis considering only Aya, EuroLLMs, Gemmas, LLama 3.1, Openchat 3.5, Qwen 2.5, and Yi. As a result, we confirmed the previous conclusion -- the \textbf{P3} prompt is the best one in this scenario in 32/36 cases. The details on this analysis can be found in Appendix, Table~\ref{tab:prompts_supporting_langs}.


\paragraph{Ranking models}



As our goal is to identify models that correct texts with the highest quality (maximizing correctness metric), preserving the meaning of the original text (maximizing metrics based on BERT Score, SentenceBERT and BLEURT), applying possibly small changes (maximizing metrics based on Levenshtein, Length difference, and GLEU), and estimating the ability not to change a given text when it is correct (maximizing $F_1$ metric), we analyzed those metrics per language and then aggregated them to obtain a general overview of these models. We left language drift aside, as it was primarily used to filter out LLMs not supporting all languages considered. 

To reach this goal, we followed a three-step analysis consisting of metric calculation, per-metric ranking creation, and global ranking creation:

(i) Metric calculation -- First, we calculated metric values for each model and each language considered. These are provided in Appendix~\ref{sec:appendix_scores}, Table~\ref{tab:appendix_scores_en} for English, and Tables~\ref{tab:appendix_scores_de},~\ref{tab:appendix_scores_it},~\ref{tab:appendix_scores_sv} for German, Italian, and Swedish, respectively. These metrics were then macro-averaged to obtain a general, language-agnostic view of these models assigning each language the same weight. The aggregated scores are presented in Table~\ref{tab:scores_average}.

(ii) Per-metric ranking creation -- For each metric considered, we ranked each model according to the metric value. We applied this procedure both to per-language scores described in Tables~\ref{tab:appendix_scores_en}-~\ref{tab:appendix_scores_sv}, and for averaged metrics from Table~\ref{tab:scores_average}. For brevity, in this paper, we explicitly present only the rankings for the aggregated metrics in Table~\ref{tab:ranks_average} as they can be easily created by sorting each language model according to a given metric.

(iii) Global ranking creation -- Finally, in order to get a single rank assigned to each model, we perform rank aggregation based on rankings introduced in the previous paragraph using the Borda method~\cite{mclean1990borda}. In the first iteration, Borda aggregation is calculated separately for semantic metrics (BERTScore, Sentence BERT, BLEURT), as well as for syntactic ones (GLEU, length diff, Levensthein). Then, we applied the Borda aggregation on LanguageTool ranks, Correct $F_1$ ranks and newly calculated semantic and syntactic ranks. This two-step scenario is required to give each perspective (text accuracy, semantics, syntax, text preservation) the same weight, when having multiple metrics for some of them (here, semantics and syntax). The results of this step are present in Table~\ref{tab:aggregated_scores}.



\paragraph{Rankings overview} The rankings presented in Table~\ref{tab:aggregated_scores} show that Gemma 9B, Qwen 2.5, and Aya are the top-ranked models when considering all languages at once. A per-language analysis shows that for English Karen models (explicitly fine-tuned for the GEC task) take the lead, with Qwen 2.5 ranked an par with Karen (strict) as the second one. Qwen and Aya are good choices for German, followed by OpenChat and Gemmas, while for Italian Gemma 9B is the best, and Aya and Qwen are the runner ups. For Swedish both Gemmas, Qwen 2.5 and Openchat work really well. Overall, taking into considerations rankings created and requirements of text quality improvement and all language support, Gemma 9B and 2B, EuroLLMs 9B and 2B as well as OpenChat and Llama 3.1 are good options. While EuroLLMs are ranked behind Aya and Qwen in some cases, they support all languages and improve grammatical correctness in each language. 


\section{Discussion}

The rankings presented in Table~\ref{tab:aggregated_scores} show that Gemma 9B is currently the best model overall for the GEC task. Even if it is not fine-tuned, it achieves outstanding results. This model can be a good foundation model for a supervised fine-tuning process. Similarly, Aya is also very good, but its limited quality on Swedish makes Gemma 9B a better choice. Qwen 2.5 is an interesting case, as it is one of the best performing models (ranked second overall, the best for German, second on English and Swedish, and third on Italian). However, it does not decrease the number of errors on Swedish introducing a slight decrease on the language tool metric. At the same time, it is scored very high on semantic and syntactic similarity metrics as well as $F_1$ score, which means that this model tends to copy Swedish texts rather than correct them. The high score of Karen models, outperforming Mistral (their base model) proves that such fine-tuning may further increase the quality of models.

\paragraph{Model size vs. quality} Four models considered are much smaller than the rest -- Phi-3 (3.7B), Gemma 2B (2B), SmolLM (1.7B), and EuroLLM (1.7B). While Phi-3's performance is mediocre, SmolLM behaves badly on the GEC task. One may think that it is due to an insufficient number of parameters. However, Gemma 2B (ranked foruth overall) and EuroLLM 1.7B (ranked 10th), which are of similar size to SmolLM (ranked 17) present outstanding performance considering their sizes. Even though Gemma 2B and EuroLLM 1.7B are very small, they understand all four languages. This proves that appropriate training (Gemma 2B is a distilled version of larger Gemma) and appropriate data (EuroLLMs are European language-focused) are more crucial than the model size itself.

\paragraph{Recurring problems}

Several recurring issues have been identified in the results produced by LLMs. The most popular examples are listed in Table~\ref{tab:recurrent_errors}. Additionally, we listed contrastive analysis of examples rated high on one metric by one model and low by another one. These examples are provided in Appendix~\ref{sec:appendix_scores}, Tables~\ref{tab:appendix_bertscore_diff}-\ref{tab:appendix_lt_differences}, representing one semantic metric, one syntactic metric, and LanguageTool. Firstly, LLMs may produce empty results (e.g., for 36\% of German examples processed with XGLM). Secondly, LLMs may generate new text instead of correcting the existing input (e.g., the majority of XGLM, SmolLM, Bloom outputs). This behavior is also occasionally observed for other models in case of short inputs provided (e.g., TowerLLM). It is a consequence of the inherent design of LLMs, which are trained on extensive corpora to predict the next token. Thirdly, LLMs sometimes provide explanations or justifications for corrections rather than simply presenting the corrected text (most frequently observed for Yi and Mistral). These models tend to be verbose and informative, even when explicitly instructed otherwise. 
An important issue observed (e.g., for Yi) is language mixing, where an actual correction in a given language is accompanied by an English comment (even if the prompt is formulated in language other than English). This behavior may be one of the main reasons for a decrease of the language drift metric values and limited languages support observed. Other kinds of recurring problems are: making paraphrases instead of small corrections, copying prompts (e.g., TowerLLM), answering in a different language (e.g., TowerLLM), and ignoring the input provided (e.g., Phi-3, frequently generating: \textit{you haven't provided a specific passage}).

\begin{table*}
  \centering
  \small
  \begin{tabular}{lcp{3cm}p{5cm}p{3cm}}
    \hline
    \textbf{Language} &
    \textbf{Model} & \textbf{Type} & \textbf{Text generated} & \textbf{Times observed in outputs} \\  
    \hline
    German & XGLM & End of sequence token & <|im\_end|> & 902 \\ \hline
    Italian & XGLM & End of sequence token & <|im\_end|> & 287 \\ \hline
    English & XGLM & End of sequence token & <|im\_end|> & 276 \\ \hline
    English & Yi & Comments added & "grammatically correct" in output text & 327 \\ \hline 
    English & Mistral & Comments added & "grammatically correct" in output text & 108 \\ \hline 
    English & Qwen & Comments added & "grammatically correct" in output text & 4 \\ \hline 
    English & Yi & Comments added & "the text is" in output text & 65 \\ \hline
    English & Tower & Comments added & "the text is" in output text & 4 \\ \hline
    English & mistral & Comments added & "no corrections needed" in output text & 279 \\ \hline 
    English & Yi & Comments added & "no corrections needed" in output text & 65 \\ \hline
    English & XGLM & Unrelated text generated & "Delete all" as the beginning of the generated text & 1420 \\ \hline
    English & SmolLM & Unrelated text generated & "the first step in writing a research paper" as the beginning of the generated text & 682 \\ \hline
    German & Bloom & Unrelated text generated + language drift &  "introduction the use of the internet has become a commonplace part" as the beginning of the generated text & 204 \\ \hline
    German & SmolLM & Unrelated text generated + language drift & "the 1960s were an era of great change in the" as output starter & 556 \\ \hline
    English & Phi-3 & no input detected & "you haven't provided a specific passage" in output text & 61 \\ \hline
    German & TowerLLM & prompt copying & "korrigieren sie den folgenden text auf rechtschreib- und grammatikfehler, nehmen sie nur minimale änderungen vor und senden sie nur den korrigierten text zurück. wenn der text bereits korrekt ist, senden sie ihn ohne erklärungen zurück" in the generated text & 36 \\ \hline
    German & TowerLLM & prompt copying + translating & "please correct the following text for spelling and grammar errors, make only minimal changes if necessary, and send back just the corrected text" in generated text & 27 \\ \hline 
    German & Yi & language mixing & "Here is the corrected" text generated before the actual correction & 30 \\ \hline

    
    
    Swedish & Gemma 2B & no input detected + Language drift & please provide the text you would like me to correct! & 9 \\ \hline
    Italian & Gemma 9B & no input detected & per favore, fornisci il testo che vuoi correggere. & 6 \\ \hline
    German & Gemma 9B & no input detected & bitte geben sie den text ein, den ich korrigieren soll. & 7 \\ \hline 
    German & Gemma 9B & no input detected & bitte geben sie mir den text zum korrekturlesen. & 3 \\ \hline 
    Swedish & Gemma 9B & no input detected & vänligen ge mig texten du vill att jag ska redigera! & 3 \\ \hline

  \end{tabular}
  \caption{Examples of recurrent problems for given models and languages.}
  \label{tab:recurrent_errors}
\end{table*}

\paragraph{Result stability} 
Due to the non-deterministic nature of LLM outputs, we also assessed the stability of the generation process to ensure that our results were consistent and that the rankings obtained in repeated runs of the experiments remained stable. For that, we selected the best-performing model (Gemma 9B) and a randomly chosen one (TowerLLM) and reran the experiments five times, maintaining all original parameters. Although minor differences were observed, their magnitudes were negligible. Across both models and all prompts, the variation in Language Tool scores did not significantly exceed 0.001. The largest difference for the Gemma 9B model was 0.00037, obtained on the Swedish dataset with prompt  \textbf{P1}. While for the TowerLLM model, it was 0.00129, obtained on the Italian dataset with prompt \textbf{P3}. Similar magnitudes of deviations were observed for other metrics. Based on these findings, we conclude that the results are both reliable and stable across runs.

\section{Karen analysis: corrections vs preservation}
\label{sec:appendix_karen}
In the field of GEC, there is an inherent dilemma: should a model prioritize making corrections or preserving the original text with minimal changes? When aiming to preserve the original text, it's crucial to consider both semantic faithfulness and syntax preservation. This challenge is particularly pronounced for LLMs, which tend to over-correct, or even paraphrase provided texts. The authors of Karen models addressed this by offering two versions: "strict" and "creative." When focusing exclusively on the English language, the Karen-strict model excels by making minimal changes and maintaining the highest similarity to the original text. It is ranked above the creative version across all metrics related to both semantic and syntactic preservation. Despite being ranked 8th in the LT score for English, the model still achieved a high score of 0.938 (with the best model scoring 0.963). Conversely, the Karen-creative model produces higher quality output (with an LT score of 0.946), but the differences between the original and corrected texts are more significant. It still maintains high semantic and syntactic preservation, especially when compared to other LLMs. This performance can be attributed to the Karen models being fine-tuned specifically for the GEC task.

\section{Conclusions}
In this paper, we analyzed whether popular LLMs not larger than 9B parameters are able to handle grammatical error correction in a multilingual manner and proposed a framework for referenceless GEC LLM comparison. We found that Gemma (2B and 9B), EuroLLM (1.7B and 9B), Openchat 3.5, and LLaMA 3.1 are able to handle all analyzed languages (EN, DE, IT, SV) with good quality. Since Gemma 9B is the top-ranked model in the multilingual scenario and the language-averaged LT score and $F_1$-correct input preservation are the highest for it,  it is the recommended model to use (answer to \textbf{RQ1}). Also, its smaller sibling -- Gemma 2B -- performs very well. We found out that some models (Gemmas, EuroLLM 9B, Qwen, Karens, Aya, and OpenChat) preserve the original text well if no error is introduced. Comparing fine-tuned models (Karen strict and Karen creative) with their base Mistral model, we see that the fine-tuned models excel in terms of overall correction quality, which shows the importance of fine-tuning to prevent hallucinations in LLMs. Thus, \textbf{RQ2} can be answered positively. Finally, we found that the longest, most concrete prompt is the best performing one overall, which answers \textbf{RQ3}. 

We made the corrections generated using all prompts and models publicly available online~\footnote{\url{https://github.com/laniqo-public/grammar-data-mtsummit25}}.

\section{Limitations} 

Our rankings focus on relatively small LLMs. However, there are much larger models trained on more data. We suppose that these may exhibit better performance, however, they are more costly, and harder to control, as most frequently they are hidden behind remote APIs. Thus, the conclusions drawn here refer to models of moderate size (up to 9B parameters),  and bigger models may perform better. Additionally, selecting LanguageTool as the source of error information may introduce some bias towards errors detected by LanguageTool. It may be that some kinds of problems are overlooked, and some false positives are produced despite the maturity of this tool. However, LT is widely used for GEC evaluation.

\bibliography{mtsummit25}

\appendix
\paragraph{Sustainability statement}
The experiments were run on a single PC machine equipped with one GeForce 4090 RTX GPU. The correction lasted about 24 hours, with additional one hour for metric evaluation. Based on these, the CO2 impact calculated using \texttt{https://mlco2.github.io/impact/} tool is equal to 3.24 kg of CO2eq.

\section{Detailed analysis and scores}
\label{sec:appendix_scores}


\begin{table*}
  \tiny
  \centering
  \begin{tabular}{lccccccccc}
    \hline
    \textbf{Prompt [lang]} & \textbf{LT$\uparrow$} & \textbf{BERT Score$\uparrow$} & 
    \textbf{SentenceBERT$\uparrow$} & \textbf{BLEURT$\uparrow$} & \textbf{Levenshtein$\uparrow$} & \textbf{Length diff$\uparrow$} & \textbf{GLEU$\uparrow$} & \textbf{Language drift$\uparrow$} & \textbf{Correct ($F_1$)$\uparrow$} \\
    \hline
P1[EN] & 0.821 & 0.779 & 0.757 & 0.545 & 0.082 & 0.514 & 0.406 & \textbf{0.067} & 0.098 \\
P2[EN] & \textbf{0.937} & 0.891 & 0.902 & 0.732 & 0.184 & \textbf{0.858} & 0.705 & 0.037 & 0.324 \\
P3[EN] & 0.923 & \textbf{0.899} & \textbf{0.904} & \textbf{0.746} & \textbf{0.236} & \textbf{0.858} & \textbf{0.724} & 0.033 & \textbf{0.448} \\ \hline 
P1[DE] & 0.755 & 0.766 & 0.751 & 0.481 & 0.044 & 0.456 & 0.380 & -0.007 & 0.066 \\
P2[DE] & 0.906 & 0.862 & 0.881 & 0.653 & 0.094 & 0.775 & 0.642 & -0.010 & 0.190 \\
P3[DE] & \textbf{0.928} & \textbf{0.888} & \textbf{0.905} & \textbf{0.701} & \textbf{0.132} & \textbf{0.848} & \textbf{0.709} & \textbf{0.001} & \textbf{0.319} \\ \hline 
P1[IT] & 0.672 & 0.760 & 0.716 & 0.447 & 0.065 & 0.449 & 0.363 & \textbf{0.042} & 0.082 \\
P2[IT] & 0.833 & 0.855 & 0.843 & 0.650 & 0.146 & 0.746 & 0.616 & 0.010 & 0.283 \\
P3[IT] & \textbf{0.906} & \textbf{0.884} & \textbf{0.887} & \textbf{0.710} & \textbf{0.183} & \textbf{0.830} & \textbf{0.686} & 0.027 & \textbf{0.372} \\ \hline 
P1[SV] & 0.630 & 0.769 & 0.755 & 0.467 & 0.045 & 0.450 & 0.366 & \textbf{0.029} & 0.068 \\
P2[SV] & 0.795 & 0.878 & 0.882 & 0.668 & 0.123 & 0.813 & 0.676 & 0.005 & 0.262 \\
P3[SV] & \textbf{0.849} & \textbf{0.904} & \textbf{0.909} & \textbf{0.719} & \textbf{0.160} & \textbf{0.883} & \textbf{0.744} & 0.009 & \textbf{0.377} \\
    \hline

  \end{tabular}
  \caption{Prompt selection vs. metrics for each language, considering only models that support all languages. For each language and prompt, a given row represents scores averaged over all 7 models supporting all four languages considered. Prompt identifiers are the same as introduced in Section~\ref{sec:prompts}.}
  \label{tab:prompts_supporting_langs}
\end{table*}

\begin{table*}
  \tiny
  \centering
  \begin{tabular}{lccccccccc}
    \hline
    \textbf{Model} & \textbf{LT (0.754)} & \textbf{BERT Score} & \textbf{SentenceBERT} & \textbf{BLEURT} & \textbf{Levenshtein} & \textbf{Length diff} & \textbf{GLEU} & \textbf{Language drift} & \textbf{Correct ($F_1$)} \\
    \hline
    \textbf{Aya} & 0.919 & 0.944 & 0.957 & 0.826 & 0.365 & 0.954 & 0.857 & 0.019 & 0.568 \\
BLOOM & 0.558 & 0.550 & 0.027 & 0.320 & 0.001 & 0.065 & 0.033 & \textbf{0.085 (1)} & 0.000 \\
\textbf{EuroLLM (1.7B)} & 0.936 & 0.810 & 0.784 & 0.594 & 0.061 & 0.802 & 0.442 & 0.051 & 0.075 \\
\textbf{EuroLLM (9B)} & 0.939 & 0.929 & 0.942 & 0.808 & 0.356 & 0.929 & 0.822 & 0.032 & 0.571 \\
\textbf{Gemma (2B)} & \textbf{0.963 (1)} & 0.924 & 0.945 & 0.766 & 0.137 & 0.935 & 0.771 & 0.027 & 0.522 \\
\textbf{Gemma (9B)} & \textbf{0.949 (3)} & 0.946 & 0.961 & 0.791 & 0.161 & 0.950 & 0.831 & 0.026 & 0.638 \\
\textbf{Karen (creative)} & 0.946 & \textbf{0.946 (3)} & \textbf{0.965 (3)} & \textbf{0.838 (3)} & \textbf{0.424 (3)} & \textbf{0.958 (3)} & \textbf{0.871 (3)}  & 0.029 & \textbf{0.646 (3)} \\
\textbf{Karen (strict)} & 0.938 & \textbf{0.956 (2)} & \textbf{0.968 (2)} & \textbf{0.857 (2)} & \textbf{0.471 (1)} & \textbf{0.959 (2)} & \textbf{0.896 (2)} & 0.015 & \textbf{0.680 (1)} \\
\textbf{Llama 3.1} & 0.939 & 0.919 & 0.928 & 0.778 & 0.250 & 0.913 & 0.792 & 0.032 & 0.429 \\
\textbf{Mistral} & \textbf{0.958 (2)} & 0.838 & 0.854 & 0.653 & 0.057 & 0.712 & 0.554 & 0.069 & 0.043 \\
\textbf{OpenChat} & 0.948 & 0.932 & 0.951 & 0.803 & 0.309 & 0.948 & 0.821 & 0.034 & 0.515 \\
Phi & 0.612 & 0.662 & 0.589 & 0.373 & 0.009 & 0.250 & 0.161 & 0.071 & 0.007 \\
\textbf{Qwen 2.5} & 0.935 & \textbf{0.961 (1)} & \textbf{0.970 (1)} & \textbf{0.861 (1)} & \textbf{0.452 (2)} & \textbf{0.960 (1)} & \textbf{0.913 (1)} & 0.004 & \textbf{0.680 (2)} \\
SmolLM & 0.233 & 0.543 & 0.045 & 0.283 & 0.001 & 0.067 & 0.032 & \textbf{0.077 (2)} & 0.000 \\
\textbf{TowerLLM} & 0.894 & 0.882 & 0.853 & 0.723 & 0.248 & 0.834 & 0.687 & 0.049 & 0.402 \\
XGLM & 0.281 & 0.535 & 0.049 & 0.229 & 0.005 & 0.149 & 0.051 & -0.096 & 0.000 \\
\textbf{Yi} & 0.774 & 0.726 & 0.699 & 0.490 & 0.036 & 0.330 & 0.271 & \textbf{0.076 (3)} & 0.032 \\ \hline

  \end{tabular}
  \caption{Scores for English. Models marked in bold improve LanguageTool score in comparison to the text before correction (0.754). Numbers (1), (2), (3) represent the top-3 ranked models on a given metric.}
  \label{tab:appendix_scores_en}
\end{table*}

\begin{table*}
  \tiny
  \centering
  \begin{tabular}{lccccccccc}
    \hline
    \textbf{Model} & \textbf{LT (0.74)} & \textbf{BERT Score} & \textbf{SentenceBERT} & \textbf{BLEURT} & \textbf{Levenshtein} & \textbf{Length diff} & \textbf{GLEU}  & \textbf{Language drift} & \textbf{Correct ($F_1$)} \\
    \hline
\textbf{Aya} & \textbf{0.971 (1)} & 0.907 & 0.938 & \textbf{0.752 (3)} & 0.169 & 0.921 & 0.769 & 0.006 & 0.376 \\
BLOOM & 0.159 & 0.508 & 0.040 & 0.218 & 0.001 & 0.055 & 0.036 & -0.978 & 0.000 \\
\textbf{EuroLLM (1.7B)} & 0.923 & 0.863 & 0.883 & 0.671 & 0.099 & 0.889 & 0.636 & 0.002 & 0.180 \\
\textbf{EuroLLM (9B)} & 0.934 & 0.888 & 0.923 & 0.706 & 0.190 & 0.803 & 0.722 & \textbf{0.009 (2)} & \textbf{0.514 (1)} \\
\textbf{Gemma (2B)} & 0.951 & \textbf{0.916 (3)} & \textbf{0.943 (3)} & 0.739 & 0.105 & 0.923 & 0.776 & 0.006 & 0.435 \\
\textbf{Gemma (9B)} & 0.940 & \textbf{0.922 (2)} & \textbf{0.949 (1)} & 0.732 & 0.097 & \textbf{0.930 (2)} & \textbf{0.787 (2)} & 0.006 & 0.351 \\
\textbf{Karen (creative)} & 0.936 & 0.913 & 0.936 & 0.708 & \textbf{0.274 (1)} & \textbf{0.930 (1)} & 0.758 & -0.164 & \textbf{0.491 (2)} \\
\textbf{Karen (strict)} & 0.945 & 0.863 & 0.882 & 0.534 & \textbf{0.222 (2)} & 0.869 & 0.580 & -0.471 & \textbf{0.455 (3)} \\
\textbf{Llama 3.1} & \textbf{0.958 (2)} & 0.898 & 0.921 & 0.730 & 0.121 & 0.890 & 0.739 & \textbf{0.014 (1)} & 0.204 \\
\textbf{Mistral} & 0.895 & 0.833 & 0.888 & 0.597 & 0.042 & 0.695 & 0.580 & -0.020 & 0.013 \\
\textbf{OpenChat} & \textbf{0.954 (3)} & 0.915 & 0.937 & \textbf{0.762 (2)} & 0.179 & 0.924 & \textbf{0.786 (3)} & \textbf{0.008 (3)} & 0.354 \\
Phi & 0.627 & 0.663 & 0.582 & 0.313 & 0.005 & 0.226 & 0.147 & -0.061 & 0.000 \\
\textbf{Qwen 2.5} & 0.940 & \textbf{0.923 (1)} & \textbf{0.948 (2)} & \textbf{0.766 (1)} & \textbf{0.205 (3)} & \textbf{0.927 (3)} & \textbf{0.816 (1)} & 0.004 & 0.449 \\
SmolLM & 0.221 & 0.539 & 0.088 & 0.242 & 0.001 & 0.066 & 0.033 & -0.978 & 0.000 \\
\textbf{TowerLLM} & 0.949 & 0.845 & 0.894 & 0.469 & 0.126 & 0.888 & 0.508 & -0.524 & 0.321 \\
XGLM & 0.699 & 0.523 & 0.083 & 0.130 & 0.006 & 0.155 & 0.043 & -0.407 & 0.000 \\
\textbf{Yi} & 0.780 & 0.757 & 0.708 & 0.448 & 0.023 & 0.421 & 0.353 & -0.051 & 0.006 \\ \hline
  \end{tabular}
  \caption{Scores for German. Models marked in bold improve LanguageTool score in comparison to the text before correction (0.74). Numbers (1), (2), (3) represent the top-3 ranked models on a given metric.}
  \label{tab:appendix_scores_de}
\end{table*}

\begin{table*}
  \tiny
  \centering
  \begin{tabular}{lccccccccc}
    \hline
    \textbf{Model} & \textbf{LT (0.851)} & \textbf{BERT Score} & \textbf{SentenceBERT} & \textbf{BLEURT} & \textbf{Levenshtein} & \textbf{Length diff} & \textbf{GLEU} & \textbf{Language drift} & \textbf{Correct ($F_1$)}  \\
    \hline
\textbf{Aya} & \textbf{0.967 (2)} & \textbf{0.917 (3)} & \textbf{0.937 (3)} & \textbf{0.784 (3)} & \textbf{0.287 (2)} & 0.921 & \textbf{0.784 (3)} & 0.022 & \textbf{0.541 (2)} \\
BLOOM & 0.008 & 0.495 & 0.028 & 0.230 & 0.001 & 0.040 & 0.025 & -0.934 & 0.000 \\
\textbf{EuroLLM (1.7B}) & 0.900 & 0.855 & 0.847 & 0.640 & 0.173 & 0.844 & 0.604 & 0.011 & 0.314 \\
\textbf{EuroLLM (9B)} & 0.889 & 0.852 & 0.851 & 0.637 & 0.156 & 0.663 & 0.590 & \textbf{0.048 (1)} & 0.305 \\
\textbf{Gemma (2B)} & \textbf{0.958 (3)} & 0.914 & 0.936 & 0.760 & 0.125 & \textbf{0.923 (3)}& 0.753 & \textbf{0.014 (3)} & 0.384 \\
\textbf{Gemma (9B)} & \textbf{0.970 (1)} & \textbf{0.938 (1)} & \textbf{0.954 (1)} & \textbf{0.803 (2)} & 0.164 & \textbf{0.944 (1)} & \textbf{0.814 (2)} & 0.029 & \textbf{0.644 (1)} \\
Karen (creative) & 0.386 & 0.829 & 0.903 & 0.436 & 0.147 & 0.870 & 0.467 & -0.645 & 0.245 \\
Karen (strict) & 0.420 & 0.847 & 0.900 & 0.480 & 0.205 & 0.884 & 0.521 & -0.607 & 0.335 \\
\textbf{Llama 3.1} & 0.924 & 0.844 & 0.833 & 0.629 & 0.115 & 0.750 & 0.577 & \textbf{0.046 (2)} & 0.167 \\
Mistral & 0.530 & 0.775 & 0.816 & 0.429 & 0.046 & 0.612 & 0.391 & -0.353 & 0.022 \\
\textbf{OpenChat} & 0.950 & 0.908 & 0.925 & 0.776 & \textbf{0.262 (3)} & 0.913 & 0.766 & 0.023 & 0.457 \\
Phi & 0.315 & 0.640 & 0.437 & 0.197 & 0.005 & 0.215 & 0.124 & -0.253 & 0.000 \\
\textbf{Qwen 2.5} & 0.912 & \textbf{0.928 (2)} & \textbf{0.946 (2)} & \textbf{0.814 (1)} & \textbf{0.291 (1)} & \textbf{0.929 (2)} & \textbf{0.819 (1)} & 0.020 & \textbf{0.518 (3)} \\
SmolLM & 0.006 & 0.525 & 0.050 & 0.247 & 0.001 & 0.051 & 0.024 & -0.934 & 0.000 \\
TowerLLM & 0.621 & 0.833 & 0.830 & 0.536 & 0.169 & 0.806 & 0.515 & -0.315 & 0.283 \\
XGLM & 0.413 & 0.478 & 0.139 & 0.097 & 0.015 & 0.230 & 0.040 & -0.653 & 0.000 \\
Yi & 0.682 & 0.801 & 0.751 & 0.549 & 0.072 & 0.584 & 0.469 & 0.027 & 0.022 \\
    \hline

  \end{tabular}
  \caption{Scores for Italian. Models marked in bold improve LanguageTool score in comparison to the text before correction (0.851). Numbers (1), (2), (3) represent the top-3 ranked models on a given metric.}
  \label{tab:appendix_scores_it}
\end{table*}

\begin{table*}
  \tiny
  \centering
  \begin{tabular}{lccccccccc}
    \hline
    \textbf{Model} & \textbf{LT (0.802)} & \textbf{BERT Score} & \textbf{SentenceBert} & \textbf{BLEURT} & \textbf{Levenshtein} & \textbf{Length diff} & \textbf{GLEU} & \textbf{Language drift} & \textbf{Correct ($F_1$)}  \\
    \hline
Aya & 0.742 & 0.910 & 0.907 & 0.724 & 0.215 & 0.917 & 0.779 & \textbf{0.014 (3)} & 0.436 \\
BLOOM & 0.010 & 0.511 & 0.019 & 0.229 & 0.001 & 0.064 & 0.035 & -0.943 & 0.000 \\
\textbf{EuroLLM (1.7B)} & 0.888 & 0.870 & 0.877 & 0.650 & 0.141 & 0.879 & 0.650 & -0.019 & 0.323 \\
\textbf{EuroLLM (9B)} & \textbf{0.899 (3)} & 0.882 & 0.894 & 0.695 & 0.172 & 0.765 & 0.675 & \textbf{0.025 (2)} & 0.360 \\
\textbf{Gemma (2B)} & 0.886 & \textbf{0.928 (3)} & \textbf{0.938 (3)} & 0.756 & 0.118 & 0.925 & 0.796 & 0.002 & 0.428 \\
\textbf{Gemma (9B)} & \textbf{0.935 (1)} & \textbf{0.942 (1)} & \textbf{0.954 (1)} & \textbf{0.771 (2)} & 0.123 & \textbf{0.944 (1)} & \textbf{0.822 (2)} & 0.005 & \textbf{0.608 (1)} \\
Karen (creative) & 0.553 & 0.893 & 0.928 & 0.591 & \textbf{0.260 (1)} & \textbf{0.937 (3)} & 0.651 & -0.407 & \textbf{0.445 (3)} \\
Karen (strict) & 0.345 & 0.849 & 0.902 & 0.428 & 0.172 & 0.912 & 0.490 & -0.666 & 0.380 \\
\textbf{Llama 3.1} & \textbf{0.926 (2)} & 0.885 & 0.896 & 0.697 & 0.106 & 0.858 & 0.691 & \textbf{0.034 (1)} & 0.218 \\
Mistral & 0.456 & 0.796 & 0.829 & 0.450 & 0.029 & 0.602 & 0.431 & -0.300 & 0.010 \\
\textbf{OpenChat} & 0.884 & 0.920 & 0.914 & \textbf{0.761 (3)} & \textbf{0.217 (3)} & 0.916 & \textbf{0.799 (3}) & 0.010 & 0.355 \\
Phi & 0.159 & 0.656 & 0.525 & 0.256 & 0.005 & 0.228 & 0.137 & -0.017 & 0.000 \\
Qwen 2.5 & 0.798 & \textbf{0.929 (2)} & \textbf{0.944 (2)} & \textbf{0.778 (1)} & \textbf{0.244 (2)} & \textbf{0.938 (2)} & \textbf{0.834 (1)} & 0.004 & \textbf{0.532 (2)} \\
SmolLM & 0.007 & 0.539 & 0.058 & 0.240 & 0.001 & 0.074 & 0.033 & -0.943 & 0.000 \\
TowerLLM & 0.146 & 0.769 & 0.782 & 0.285 & 0.019 & 0.688 & 0.290 & -0.744 & 0.000 \\
XGLM & 0.069 & 0.516 & 0.110 & 0.215 & 0.001 & 0.136 & 0.043 & -0.875 & 0.000 \\
Yi & 0.680 & 0.867 & 0.855 & 0.640 & 0.100 & 0.809 & 0.650 & 0.010 & 0.128 \\
    \hline

  \end{tabular}
  \caption{Scores for Swedish. Models marked in bold improve LanguageTool score in comparison to the text before correction (0.802). Numbers (1), (2), (3) represent the top-3 ranked models on a given metric.}
  \label{tab:appendix_scores_sv}
\end{table*}

\begin{table*}
  \centering
  \small
  \begin{tabular}{lp{110mm}cc}
    \hline
    \textbf{Source} & \textbf{Text} & \textbf{Score} \\ 
    \hline
    Original text & It is wonderful to hear that there is a good variety of activities available at the camp. & - \\ \hline 
    Gemma & It is wonderful to hear that there is a good variety of activities available at the camp. & 1.0 \\ \hline 
    TowerLLM & There are no errors in this sentence. & 0.73 \\ \hline \hline
    Original text & Subject: About daily life at our school & - \\ \hline 
    Aya & Subject: About daily life at our school & 1.0 \\ \hline 
    TowerLLM & The edited text is as follows:

Subject: About daily life at our school 

Dear Parents/Guardians,

I am writing to provide you with an update on your child's daily life at our school. We strive to create a safe and nurturing environment where students can learn and grow academically, socially, and emotionally. Our teachers work tirelessly to ensure that each student receives individualized attention and support in achieving their full potential. Here are some highlights of what we offer: (...) & 0.65 \\ \hline \hline
    Original text & IN CASE YOU DECIDE TO CHANGE THE PROGRAM, WE SUGGEST ON TUSDAY GO TO THE SHOW AND ON WENSDAY INSTED FREE TIME, VISIT THE SCIENCE MUSEUM. & - \\ \hline
    Aya & IN CASE YOU DECIDE TO CHANGE THE PROGRAM, WE SUGGEST ON TUESDAY GO TO THE SHOW AND ON WEDNESDAY INSTEAD OF FREE TIME, VISIT THE SCIENCE MUSEUM. & 0.97 \\ \hline
    Gemma & In case you decide to change the program, we suggest on Tuesday go to the show and on Wednesday instead of free time, visit the science museum. & 0.55 \\ \hline \hline 
    Original text: & For me was this aventure one of the Best. & - \\ \hline 
    Aya & For me, this adventure was one of the best. & 0.88 \\ \hline
    TowerLLM & There were no errors in the original text. The text "For me wa" -- which should be "for me was," has been changed to its correct form. & 0.6 \\ \hline \hline 
    
    Orignal text & You offered something impossible to do. & - \\ \hline
    Aya & You offered something impossible to do. & 1.0 \\ \hline 
    TowerLLM & There were no spelling or grammatical errors in the original text. The AI did not add any additional information beyond correction of the existing text. & 0.64 \\ \hline \hline 
    Original text & I look forward to hearing from you soon. & - \\ \hline 
    Aya & I look forward to hearing from you soon. & 1.0 \\ \hline 
    TowerLLM  & Here is a possible response:

I am glad that my message was helpful! Thank you for reaching out to me. I would be happy to assist you further with your inquiry or concern. Please let me know if there's anything else I can do for you. Best regards, [Your Name] & 0.66 \\ \hline
    
  \end{tabular}
  \caption{Examples of large differences between BERT scores measured on pairs of models for given input texts. The scores are calculated using  the \texttt{bert-base-multilingual-cased} model Each example is represented as a triple: original text, model scoring high, model scoring low.}
  \label{tab:appendix_bertscore_diff}
\end{table*}

\begin{table*}
  \centering
  \small
  \begin{tabular}{lp{110mm}}
    \hline
    \textbf{Source} & \textbf{Text} \\ 
    \hline

    Original text & Helen Ryan \\ \hline 
    Aya & Helen Ryan  \\ \hline
    Tower & Here's the corrected text:
Helen Ryan edited the following text for spelling and grammar errors, made minor adjustments, and returned only the corrected version of the text. If the original text w
as already perfect, she simply sent back a confirmation message with no edits or corrections.  \\ \hline \hline 

    Original text & Thank you.  \\ \hline 
    Openchat & Thank you.  \\ \hline
    Gemma & Thank you. $\backslash$n$\backslash$n$\backslash$nLet me know if you have anything else I can help with!  \\ \hline \hline 

    Original text & Yeah!! \\ \hline 
    Aya & Yeah!!  \\ \hline
    Gemma & Yay!  \\ \hline \hline 

    Original text & 13/06/00  \\ \hline 
    TowerLLM & 13/06/00  \\ \hline 
    OpenChat & The date provided appears to be a mix of numbers and letters, which doesn't seem like a standard format for a date or a sentence. Please provide
a clearer context or a different input so I can assist you with editing for spelling and grammar mistakes  \\ \hline \hline

Original text & I'm fine. \\ \hline 
Aya & I'm fine. \\ \hline 
TowerLLM & Sure! Here's your edited text: $\backslash$n
I'm fien.  \\ \hline \hline 

    Original text & love  \\ \hline 
    Openchat & Love  \\ \hline 
    TowerLLM & The following text has no errors or improvements needed. Here is your original text: I am a student who loves to learn new things every day. I have always been interested in science and technology since my childhood days. My passion for learning led me to pursue a degree in computer engineering at university. During my studies, I discovered that programming was one of the most fascinating aspects of computing. (...)  \\ \hline \hline

    Original text & This machines are connected to Internet, and it gives to my family too much facility to get every information we want, from how to prepare son drink to what was the result of the Sunday's football match. \\ \hline 
    Aya & This machines are connected to internet, and it gives to my family too much facility to get every information we want, from how to prepare son drink to what was the result of the Sunday's football match. \\ \hline 
    TowerLLM & These machines are connected to the internet, which provides my family with great convenience in obtaining all kinds of information they need, such as how to prepare a certain beverage or checking the results of last weekend's football game. \\ \hline 
    
  \end{tabular}
  \caption{Examples of big differences between Levenshtein scores for given input texts. Each example is represented as a triple: original text, model scoring high, model scoring low.}
  \label{tab:appendix_levenshtein_diff}
\end{table*}

\begin{table*}
  \centering
  \small
  \begin{tabular}{lp{80mm}p{45mm}}
    \hline
    \textbf{Source} & \textbf{Text} & \textbf{Errors} \\
    \hline

    Original text & I was thinking 'what should I do?' I haven't got my mobile phone with me And I was half way between my school and home. & possible missing comma after "thinking", half way -> halfway \\ \hline 

    Aya & I was thinking, "What should I do?" I hadn't brought my mobile phone with me, and I was halfway between my school and home. & - \\ \hline

    TowerLLM & "I" -- minor change made to capitalize first person pronoun
"'" -- added apostrophe to indicate a thought bubble or speech balloon in comics/manga style writing
"And" -- added conjunction to connect two independent clauses
"half way" -- changed "hal"f wa"y" to "halfway" as per standard English usage
"my schoo"l" -- changed "schoo"l" to "school" as per standard English usage
"home" -- changed "hom"e" to "home" as per standard English usage & Several (12) mistakes due to wrong/extra apostrophes added. \\ \hline \hline 

Original text &	Propably in the future we might be to busy to go to our own home and spend some time there. & Propably - spelling, to busy -> too busy \\ \hline 
Aya &	Probably in the future we might be too busy to go to our own home and spend some time there. & - \\ \hline 
TowerLLM &	Probablemente en el futuro podríamos estar demasiado ocupados para ir a nuestra propia casa y pasar un tiempo allí. & Multiple errors due to language switch \\ \hline \hline 

Original text & Today, however, people do not need to spend many times on making food, washing up, or even on turning on an off the light. & No errors detected \\ \hline 
Gemma & Today, however, people do not need to spend much time on making food, washing up, or even on turning a light on and off. & - \\ \hline 

Aya & Today , however , people do not need to spend much time on cooking , cleaning up , or even switching lights on and off . & Multiple errors due to extra whitespaces before punctuation marks.\\ \hline \hline

Original text & It became scencial to do my homework. & scencial - spelling mistake \\ \hline 
Aya & It became essential to do my homework. & - \\ \hline 
TowerLLM & The text "It becamE scencial tO dO mY homEworK." is incorrect. Please provide a revised version with the intended correction(s). & Multiple errors due to capitalization \\ \hline \hline 

Original text &	It was writen too that there were discounts available and I asked for them and they said me that they were n't allow to do any discount. & writen -> speling , were n't - mistake, missing commas before "and"s, "said me" missing preposition, "too that" -> "to that" suggested\\ \hline 
Openchat & It was written that there were discounts available, and I asked for them, and they said they weren't allowed to give any discount. & - \\ \hline 
Aya & It was written too that there were discounts available and I asked for them and they said me that they weren't allow to do any discount. & missing commas before "and"s, "said me" missing preposition, "too that" suggested to change with "to that" \\ \hline 

  \end{tabular}
  \caption{Examples of high differences between LanguageTool scores for given input texts. Each example is represented as a triple: original text, model scoring high, model scoring low.}
  \label{tab:appendix_lt_differences}
\end{table*}

\end{document}